\pdfoutput=1

\documentclass[11pt]{article}

\usepackage{acl}

\usepackage{times}
\usepackage{latexsym}

\usepackage{pifont}
\usepackage{amssymb}
\usepackage{pgfplots}
\usepackage{pgfplotstable}
\usepackage{adjustbox}
\usepackage{caption}
\usepackage{subcaption}
\pgfplotsset{compat=newest}
\usetikzlibrary{matrix,backgrounds}
\usepackage{siunitx}

\pgfplotsset{compat=1.18}

\usepackage{tcolorbox}

\usepackage[table]{xcolor}
\usepackage{graphicx}
\usepackage{colortbl}
\usepackage{adjustbox}
\usepackage{booktabs}
\usepackage{multirow}
\usepackage[T1]{fontenc}

\usepackage[utf8]{inputenc}

\usepackage{microtype}

\usepackage{inconsolata}

\usepackage{graphicx}

\tcbuselibrary{breakable}
\tcbuselibrary{skins}

\title{Effects of Cross-lingual Evidence in Multilingual Medical Question Answering}

 \author{Anar Yeginbergen \and Maite Oronoz \and Rodrigo Agerri \\
         HiTZ Center - Ixa, University of the Basque Country UPV/EHU\\
         \texttt{\{anar.yeginbergen,maite.oronoz,rodrigo.agerri\}@ehu.eus}
         }

\definecolor{lightblue}{RGB}{220, 230, 241}
\definecolor{blue}{RGB}{184, 204, 228}
\definecolor{lightgreen}{RGB}{200, 250, 237}
\definecolor{darkgreen}{RGB}{70, 212, 178}
\definecolor{lightred}{RGB}{244, 204, 204}
\definecolor{orange}{RGB}{252, 189, 126}
\definecolor{darkred}{RGB}{224, 102, 102}
\definecolor{gray}{RGB}{242, 242, 242}
\definecolor{headercolor}{RGB}{255, 255, 153}
\definecolor{highlight}{RGB}{153, 255, 255}

\begin{document}
\maketitle

\begin{abstract}

 This paper investigates Multilingual Medical Question Answering across high-resource (English, Spanish, French, Italian) and low-resource (Basque, Kazakh) languages. We evaluate three types of external evidence sources across models of varying size: curated repositories of specialized medical knowledge, web-retrieved content, and explanations from LLM's parametric knowledge. Moreover, we conduct experiments with multilingual, monolingual and cross-lingual retrieval. Our results demonstrate that larger models consistently achieve superior performance in English across baseline evaluations. When incorporating external knowledge, web-retrieved data in English proves most beneficial for high-resource languages. Conversely, for low-resource languages, the most effective strategy combines retrieval in both English and the target language, achieving comparable accuracy to high-resource language results.
 These findings challenge the assumption that external knowledge systematically improves performance and reveal that effective strategies depend on both the source of language resources and on model scale.  Furthermore, specialized medical knowledge sources such as PubMed are limited: while they provide authoritative expert knowledge, they lack adequate multilingual coverage
 . Code and resources publicly available: 
 \url{https://github.com/anaryegen/multilingual-medical-qa/}

\end{abstract}

\section{Introduction}
The rapid advancements in Large Language Models (LLM) research have yielded impressive results across various domains, including healthcare \citep{brown2020language, achiam2023gpt, lievin2024can}. LLMs demonstrate strong capabilities in clinical reasoning and decision-making across tasks of varying complexity, opening the door to potential applications in real-world medical contexts \cite{chen2024huatuogpto1medicalcomplexreasoning,sellergren2025medgemma, shool2025systematic}.

\begin{figure}[t]
    \centering
    \includegraphics[width=1.0\linewidth]{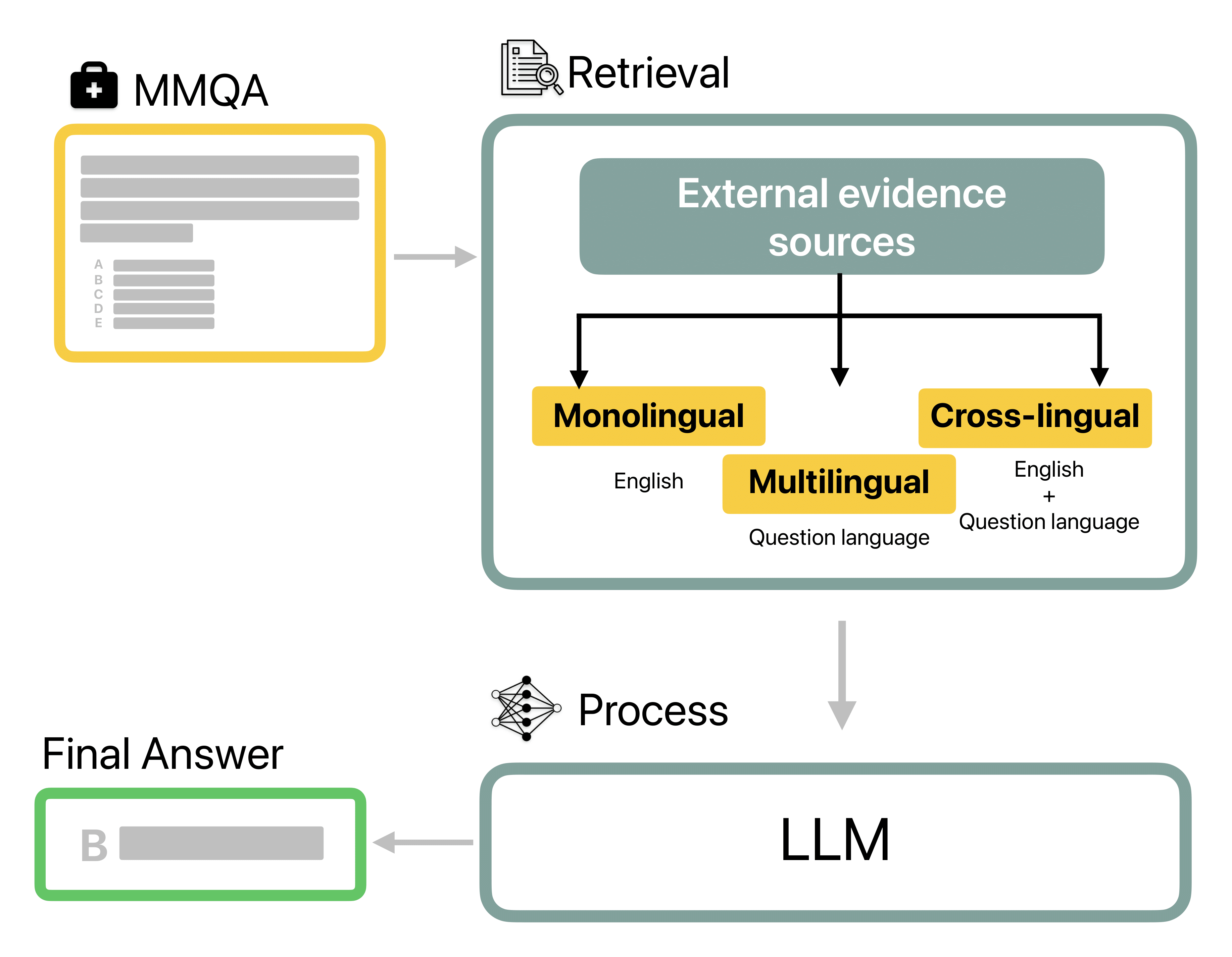}
    \caption{Illustration of the Multilingual (English, Spanish, Italian, French, Kazakh, and Basque) Medical Question Answering pipeline. We obtain external knowledge from: i) the web, ii) parametric knowledge of LLMs, and iii) retrieved passages from the knowledge sources such as PubMed, Wikipedia, and other popular medical data sources. Finally, we provide these documents to some state-of-the-art LLMs and generate the answer.}
    \label{fig:intro_img}
\end{figure}

Nevertheless, despite the continuously increasing functionalities of LLMs, they still struggle with hallucinations, limited context length, and ungrounded generation, which undermine their consistency and factual accuracy, potentially causing serious harm in critical domains such as medicine \citep{ahmad2023creating, yu2023leveraging, kim2025medical, jcm14176169, roustan2025clinicians}. 


To address these challenges, several strategies have been explored: including training specialized medical models \citep{sellergren2025medgemma, huatuogpt-2023, chen2024huatuogpto1medicalcomplexreasoning}, curating more diverse datasets \citep{jin2021disease, jin2019pubmedqa, pal2022medmcqa}, and grounding outputs in established knowledge sources \citep{xiong-etal-2024-benchmarking, ALONSO2024102938, biesheuvel2025large}. Among these, relying on established knowledge sources through methods such as Retrieval-Augmented Generation (RAG) has become especially prominent, as it helps compensate for the gaps in LLMs’ internal knowledge. 

In any case, most of this work is concentrated on English, making it difficult to generalize previous research findings across other languages. For instance, \citet{ALONSO2024102938} highlight a stark performance drop for French, Italian, and Spanish compared to English, even when using RAG \cite{xiong-etal-2024-benchmarking}, indicating that non-English medical Question Answering (QA) remains under-researched. Additionally, almost all the curated databases with medical expert knowledge are in English \citep{xiong-etal-2024-benchmarking, amugongo2025retrieval}, which makes it more challenging to deal with medical exams in other languages.




At the same time, research has demonstrated that scaling up LLMs enables them to encode substantial amounts of domain-specific information in their parametric knowledge \citep{brown2020language, ren2023investigating}. This highlights the need to better understand how to combine LLMs' parametric knowledge pertaining to the medical domain with automatically retrieved external information to achieve optimal performance in Medical QA. Taking this into consideration, we formulate the following research questions:
\begin{itemize}
    \item Is there a universally effective knowledge-augmentation method for multilingual clinical QA?
    \item Does retrieval method performance vary systematically across languages, or does one approach consistently outperform others regardless of the target language?
    \item Do we obtain better results in medical QA by retrieving the external knowledge from recognized authoritative medical sources, such as PubMed, or by directly querying the Web? 
    \item How does answer accuracy differ when retrieval and generation are performed monolingually in English, multilingually in the language of the question, and cross-lingually to compensate for the lack of resources in some languages?
    \item Can current LLMs perform well without external retrieval, or is domain-specific knowledge augmentation essential for reliable performance across languages? 
\end{itemize}

Our experimental results reveal the following insights. First, retrieving evidence from English web search and providing it to LLMs yields the best performance for MMQA across high-resource languages, and combining English retrieval with the target language retrieval is the best strategy for low-resource languages. For models under 30B parameters, this English web-search strategy provides the greatest benefit, achieving an 8.2\%\ improvement over the baseline. Second, larger models consistently outperform smaller models across all evaluation settings. However, for models of 70B parameters (or larger), incorporating retrieved evidence decreases performance by 2.4\%\ on average, suggesting these models have already internalized relevant medical knowledge during pre-training. Third, and most importantly, the performance gap between high-resource and low-resource languages seems to be due to the heavy under-representation of low-resource languages during LLM pretraining, which can be compensated for by augmenting with English and target language evidence. 
Fourth, although the retrieved evidence helps smaller models to improve their accuracy, it still underperforms compared to the baselines, with the largest LLMs remaining much better in the task. 
Lastly, curated and traditionally reliable repositories of medical information exhibit less information in the medical domain compared to the Web. In English web-retrieved data, fewer than 20\% of documents came from well-known sources such as PubMed and Wikipedia, with the remainder from other medical websites. For other languages, this number drops below 0.03\%, this highlights a severe lack of expert-curated medical content in non-English data storage.

Our results challenge a common assumption in knowledge-augmented medical Question Answering: that adding external evidence systematically improves performance \cite{xiong-etal-2024-benchmarking, shi2025mkrag, sohn2025rationale}. Instead, our multilingual experiments highlight complex dynamics that vary across languages and model sizes, showing that retrieval strategies cannot be one-size-fits-all. To investigate these dynamics, we examine three types of external evidence: (1) medical local knowledge repositories such as PubMed and Wikipedia, (2) web-retrieved documents, and (3) LLM-generated evidence. Our approach is illustrated in Figure \ref{fig:intro_img}. 

Our main objective is to establish how external information obtained from sources such as web-based retrieval, curated medical repositories or LLMs' parametric knowledge interact with models ranging from 7B to 70B+ parameters across MMQA tasks, reporting detailed findings on their relative performance. This will allow us to conclude which strategy is better to perform MMQA, especially when low resourced language are involved. Our contributions are the following:

This paper provides a systematic empirical analysis of different external evidence integration strategies for Multilingual Medical Question Answering (MMQA) across languages with varying resource availability and LLMs of different sizes. Our contributions are the following: 
\begin{itemize}
\item \textbf{English-centric retrieval dominates multilingual medical QA.}
Across six languages and all tested models, retrieving evidence from English web sources consistently outperforms retrieval in the language of the question for high-resource languages. This highlights the combined impact of English-heavy medical content on the Web, on the databases of medical knowledge, and English-biased pretraining of current LLMs.

\item \textbf{Cross-lingual retrieval is essential for low-resource languages.}
For Basque and Kazakh, we show that combining English and target-language evidence substantially improves accuracy and closes the performance gap with higher-resource languages, whereas monolingual retrieval alone is insufficient.
\item \textbf{Retrieval benefits depend strongly on model scale and evidence source.}
While smaller and mid-sized models benefit from external evidence, models with parameters more than 70B often experience performance degradation when augmented with additional context, suggesting that these models have already internalized sufficient medical domain knowledge during pre-training on large-scale, diverse data, which leads to the knowledge conflict between external non-parametric knowledge and the knowledge encoded in the LLMs' parameters.

\end{itemize}

\begin{figure*}
    \centering
    \includegraphics[width=1\linewidth]{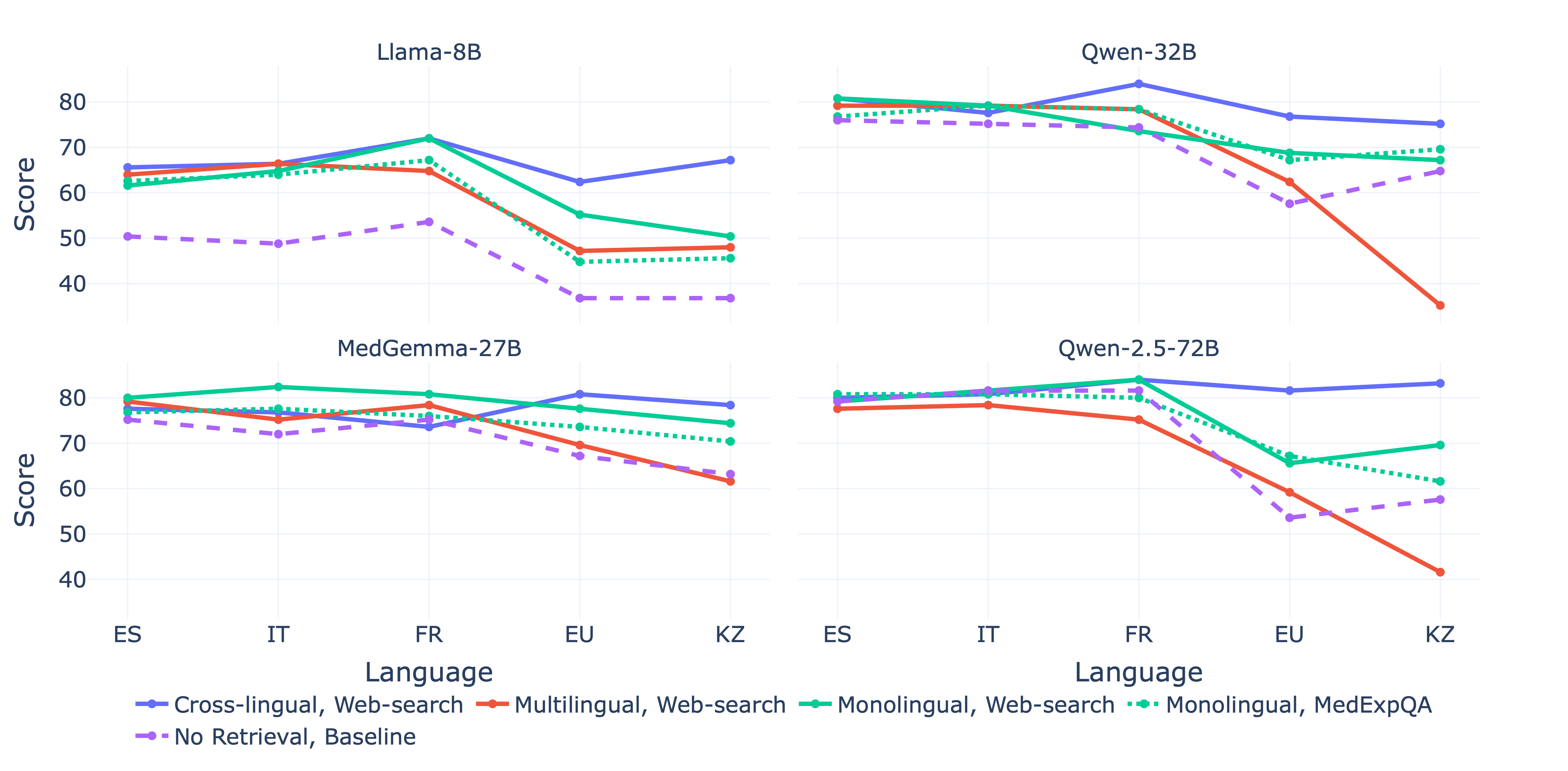}
    \caption{The comparison between different retrieval settings: monolingual, multilingual and cross-lingual across different models.}
    \label{fig:retrieval_comparison}
\end{figure*}

\section{Related work}

The application of  LLMs in the medical domain is one of the crucial directions in the development of AI. The growth of model sizes, along with their parameter counts, suggests that current language models encode a significant amount of specialized knowledge \citep{wei2022emergent,  Singhal2022LargeLMA}. This capability has sparked extensive research into evaluating how well state-of-the-art LLMs perform on specialized clinical tasks and medical knowledge domains \cite{huatuogpt-2023, achiam2023gpt, bai2023qwen, xiong-etal-2024-benchmarking, sellergren2025medgemma}. 

Recent research has systematically assessed the medical competency of foundation models, including Llama \citep{touvron2023llama, grattafiori2024llama}, Mistral \citep{jiang2024mistral}, Gemma \citep{team2025gemma}, and Qwen \citep{bai2023qwen}, among others. 
While Large Language Models (LLMs) exhibit strong performance in medical knowledge recall and clinical reasoning \cite{achiam2023gpt, xiong-etal-2024-benchmarking}, significant challenges persist in ensuring reliability, detecting hallucinations, and maintaining consistency with established clinical guidelines \cite{huatuogpt-2023, wu2025medreason}.

In the remainder of this section, we focus on reviewing prior work on medical question answering, including datasets, systems, and techniques for retrieving and applying external knowledge sources and reasoning mechanisms.

\textbf{Medical QA}. Several datasets have been constructed with the explicit aim of evaluating these limitations. \citet{jin2019pubmedqa} introduced PubMedQA, a biomedical question-answering dataset.
\citet{jin2021disease} introduced MedQA
a multiple-choice question benchmark derived from real medical licensing exams.
\citet{pal2022medmcqa} designed MedMCQA, a medical multiple-choice question answering dataset written in English and specifically focused on real medical entrance exam questions in India.

Originally in Spanish, \citet{Agerri2023HiTZAntidoteAE} created a parallel multilingual dataset from Spanish Resident Medical Intern exams. 
The dataset was used to create the first multilingual benchmark for medical QA enriched with RAG techniques.

\textbf{RAG in the medical domain.} To address the issues of knowledge scarcity in some medical areas and questions, methods such as Retrieval Augmented Generation (RAG) have been explored \citep{lewis2020retrieval, gargari2025enhancing}. For instance, the retrieval strategy  MedRAG \citep{xiong-etal-2024-benchmarking} comes as a part of a framework for medical RAG. MKRAG \citep{shi2025mkrag} incorporates fact-based retrieval from external medical knowledge bases and demonstrates a 4\% improvement in accuracy on the MedQA benchmark. MedExpQA \citep{ALONSO2024102938} is a benchmark designed to evaluate medical question-answering systems that leverage external knowledge sources, with a focus on multilingual capabilities. \citet{xiong2024improving} proposed a retrieval system that iteratively improves search results by incorporating continuous follow-up questions. 
\citet{kim2025rethinking} provided a comprehensive evaluation of RAG on different medical tasks.

\section{Sources of Medical Information}
\label{sec: sources}
This section details the methodology adopted to answer our research questions. We aim to identify the optimal strategy for selecting external knowledge sources for MMQA and analyze how different sources influence answer quality: curated medical knowledge repositories (Section \ref{sec:medexp}), retrieved from the Web (Section \ref{sec:web_search}), or the parametric knowledge encoded in LLMs (Section \ref{sec:parametric}).



\subsection{Retrieving From Medical Knowledge Sources}
\label{sec:medexp}

In the MedExpQA benchmark \citep{ALONSO2024102938}, a collection of 32 related documents was retrieved using MedRAG method \citet{xiong-etal-2024-benchmarking} from multiple knowledge sources for each question, including PubMed, Wikipedia, StatPearls, and medical textbooks. As the text stored in these repositories is predominantly in English\footnote{\url{https://meta.wikimedia.org/wiki/List_of_Wikipedias_by_language_group}} \cite{hamad2024medical}. After manual examination, were determined that the most relevant and informative documents are in this language. Hence, in our experiments, we used the English subset of the retrieved documents. The retrieved 32 documents are the result of a combination of using BM25 \citep{robertson2009probabilistic} and MedCPT \citep{jin2023medcpt} ranked by relevance, where the highest-ranked document corresponds to the highest similarity score with respect to the clinical case question. 

\citet{xiong-etal-2024-benchmarking} outlined that 32 documents is the optimal number of documents for RAG settings for the medical domain. We repeated the experiments by including top \textit{\{1, 3, 5, 10, 20, 32\}} documents to the model. The results in Figure \ref{fig:medexpqa-doccount-main} show that for the majority of the latest models, 10 documents is the most beneficial strategy, while adding more documents results in marginal gains or even performance degradation. Henceforth, we use the top 10 retrieved documents in our experiments.


\subsection{Web Search}
\label{sec:web_search}

As we established that 10 documents is the most beneficial strategy, we aimed to retrieve 10 documents from the other resources. First, we generate search queries, as described in Appendix \ref{appendix:search-query-generation}.
We retrieve documents from the web using two APIs: the web search tool by Cohere API\footnote{https://cohere.com/} and Google Search using Serper API\footnote{https://serper.dev/}. 
Although the former provides both retrieved documents and a summary generated by its underlying language model in response to the query, resulting in a more comprehensive answer, we do not include generated summaries in our experiments for fair comparison.

The imbalance of the information available on the web for less-resourced languages is evident, and we report the ratio of the retrieved data per language in Table \ref{tab:web_retrieval_comparison}. 


\begin{table*}[h]
\scriptsize 
\setlength{\tabcolsep}{0.8pt} 
\renewcommand{\arraystretch}{1} 


\begin{center}
\begin{tabular}{c|cc|cc|ccc|cc>{\columncolor{lightgreen}}c}
\rowcolor{headercolor}
\textbf{Params} & \multicolumn{2}{c|}{\textbf{8B}} & \multicolumn{2}{c|}{\textbf{12B-14B}} & \multicolumn{3}{c|}{\textbf{27-32B}} & \multicolumn{2}{c|}{\textbf{>=70B}} & \textbf{Avg (Std.)} \\
\rowcolor{headercolor}
\hline
Models & LLaMA & Qwen & Gemma & Qwen & Gemma & MedGemma & Qwen & LlaMA & Qwen & \\
\rowcolor{headercolor}
& \textbf{8B} & \textbf{8B} & \textbf{12B} & \textbf{14B} & \textbf{27B} & \textbf{27B} & \textbf{32B} & \textbf{70B} & \textbf{72B} & \\
\hline
\rowcolor{orange}
\multicolumn{11}{c}{\textbf{Baseline: no-retrieval}} \\
EN  & 61.6 & 69.6 & 63.2 & 72.0 & 70.4 & 76.8 & 80.8 & 76.0 & \underline{\textbf{84.0}} & 72.7 (7.07) \\
ES  & 50.4 & 65.6 & 60.8 & 64.8 & 72.8 & 75.2 & 76.0 & \underline{80.0} & 79.2 & 69.4 (9.24) \\
FR  & 53.6 & 58.4 & 62.4 & 67.2 & 73.6 & 75.2 & 74.4 & \underline{\textbf{83.2}} & 81.6 & 69.9 (9.65) \\
IT  & 48.8 & 62.4 & 68.8 & 67.2 & 68.0 & 72.0 & 75.2 & 79.2 & \underline{81.6} & 69.2 (9.22) \\
EU  & 36.8 & 45.6 & 48.0 & 57.6 & 62.4 & 67.2 & 57.6 & \underline{71.2} & 53.6 & 55.6 (10.26) \\
KK  & 36.8 & 44.0 & 54.4 & 42.4 & 53.6 & 63.2 & 64.8 & \underline{72.8} & 57.6 & 54.4 (11.04) \\
\rowcolor{lightgreen}
Avg. (Std.) & 48.0 (8.88) & 57.6 (9.67) & 59.6 (6.69) & 61.87 (9.71) & 66.8 (6.96) & 71.6 (4.88) & 71.47 (7.82) & 77.07 (4.17) & 72.93 (12.39) & \\
\hline
\rowcolor{orange}
\multicolumn{11}{c}{\textbf{MedExpQA}} \\
EN & 72.8 & 77.6 & 67.2 & 76.8 & 74.4 & 70.4\textcolor{red}{$\downarrow$} & 80.0\textcolor{red}{$\downarrow$} & 79.2 & \underline{81.6}\textcolor{red}{$\downarrow$}  & 75.56 (4.48) \\
ES & 62.6 & 68.8 & 70.4 & 72.0 & 79.2 & 76.8 & 76.8 & 80.0= & \underline{80.8} & 75.16 (5.76) \\
FR & 67.2 & 68.8 & 68.0 & 75.2 & 78.4 & 76.8 & 78.4 & 79.2\textcolor{red}{$\downarrow$} & \underline{80.0}\textcolor{red}{$\downarrow$} & 74.58 (4.87) \\
IT & 64.0 & 70.4 & 63.2\textcolor{red}{$\downarrow$} & 73.6 & 75.2 & 77.6 & 79.2 & \underline{81.6} & 80.8\textcolor{red}{$\downarrow$} & 73.96 (6.46) \\
EU & 44.8 & 59.2 & 63.2 & 58.4 & 67.2 & \underline{73.6} & 67.2 & \underline{73.6} &  67.2 & 63.82 (8.42) \\
KK & 45.6 & 56.0 & 56.0 & 61.6 & 70.4 & 70.4 & 69.6 & \underline{71.2\textcolor{red}{$\downarrow$}} & 61.6 & 62.49 (8.32) \\
\rowcolor{lightgreen}
Avg. & 59.5 (10.61) & 66.8 (7.21) & 64.67 (4.65) & 69.6 (7.01) & 74.13 (4.22) & 74.13 (2.91) & 75.2 (4.95) & 77.47 (3.74) & 75.33 (7.91) & \\
\hline
\rowcolor{orange}
\multicolumn{11}{c}{\textbf{LLM's parametric knowledge}} \\
EN & 67.2 & 73.6 & 69.6 & 75.2 & 67.2\textcolor{red}{$\downarrow$} & 68.8\textcolor{red}{$\downarrow$} & 75.2\textcolor{red}{$\downarrow$} & 76.0= & \underline{80.0}\textcolor{red}{$\downarrow$} & 72.53 (4.25) \\
ES & 63.2 & 70.4 & 66.4 & 74.4 & 65.6\textcolor{red}{$\downarrow$} & 74.4\textcolor{red}{$\downarrow$} & 74.4\textcolor{red}{$\downarrow$} & \underline{79.2}\textcolor{red}{$\downarrow$} & 78.4\textcolor{red}{$\downarrow$} & 71.82 (5.39) \\
FR & 68.0 & 70.4 & 60.0\textcolor{red}{$\downarrow$} & 73.6 & 64.8\textcolor{red}{$\downarrow$} & 72.8\textcolor{red}{$\downarrow$} & 75.2 & 76.8\textcolor{red}{$\downarrow$} & \underline{80.0}\textcolor{red}{$\downarrow$} & 71.29 (5.87) \\
IT & 68.0 & 71.2 & 66.4\textcolor{red}{$\downarrow$} & 78.4 & 64.8\textcolor{red}{$\downarrow$} & 74.4 & 79.2 & 80.0 & \underline{80.8}\textcolor{red}{$\downarrow$} & 73.69 (5.91) \\
EU & 53.6 & 60.8 & 65.6 & 62.4 & 65.6 & 71.2 & 71.2 & \underline{76.0} & 69.6 & 66.22 (6.33) \\
KK & 61.6 & 58.4 & 58.4 & 59.2 & 66.4 & 66.4 & \underline{72.0} & \underline{72.0}\textcolor{red}{$\downarrow$} & 63.2 & 64.18 (5.07) \\
\rowcolor{lightgreen}
Avg. (Std.) & 63.6 (5.09) & 67.47 (5.71) & 64.4 (3.91) & 70.53 (7.10) & 65.73 (0.85) & 71.33 (2.94) & 74.53 (2.59) & 76.67 (2.59) & 75.33 (6.62) & \\
\hline
\rowcolor{orange}
\multicolumn{11}{c}{\textbf{ \textit{Web Search}}} \\
EN & 72.0 & 75.2 & 72.8 & 80.0 & 73.6 & 72\textcolor{red}{$\downarrow$} & 76.8\textcolor{red}{$\downarrow$} & 79.2 & \underline{82.4}\textcolor{red}{$\downarrow$} & 76.0 (3.60) \\
ES & 61.6 & 73.6 & 75.2 & 78.4 & 84 & 80.0 & 80.8 & 82.4 & 79.2 & 77.24 (6.32) \\
FR & 72 & 72 & 76.8 & 79.2 & 81.6 & 80.8 & 73.6 & 82.4\textcolor{red}{$\downarrow$} & \underline{84.0} & 78.04 (4.35) \\
IT & 64.8 & 71.2 & 74.4 & 80.0 & 83.2 & 82.4 & 79.2 & 82.4 & \underline{\textbf{81.6}} & 77.69 (5.93) \\
EU & 55.2 & 60.8 & 71.2 & 65.6 & 75.2 & \underline{\textbf{77.6}} & 68.8 & 73.6 & 65.6 & 68.18 (6.78) \\
KK & 50.4 & 60.8 & 68.0 & 65.6 & 73.6 & \underline{\textbf{74.4}} & 67.2 & 72.8 & 69.6 & 66.93 (7.12) \\
\rowcolor{lightgreen}
Avg. (Std.) & \textbf{62.7} (8.02) & \textbf{68.93} (5.89) & 73.07 (2.87) & \textbf{74.8} (6.53) & \textbf{78.53} (4.49) & \textbf{77.87} (3.66) & \textbf{74.4} (5.06) & \textbf{78.8 }(4.12) & \textbf{77.07} (6.94) & \\
\hline

\end{tabular}

\caption{Performance comparison across different monolingual English retrieval methods, model sizes and languages. The models are grouped by the parameter size. The \underline{\textbf{underlined and bold results}} show the best results per language (row), and \textbf{bold text} means the best performing model per parameter count (column). The \textcolor{red}{$\downarrow$}  indicates the drop in performance and = indicates no improvement compared to the baseline. \underline{Underlined results} correspond to the best results for each language in a given retrieval setting.}  
\label{tab:main_results}

\end{center}
\end{table*}


\subsection{Parametric knowledge of LLMs}
\label{sec:parametric}

Our initial results show that the performance of the LLMs with more than 70B parameters excels in medical QA in English without any additional information, which leads us to hypothesize that these LLMs pre-trained on an immense amount of world knowledge encode sufficient information in the medical domain within their parameters \cite{allen2023physics, ju2024large, he-etal-2025-language}. 
Therefore, to test this hypothesis, we ask LLMs to generate answers to the queries used for web search and generate an answer for the clinical case question. 
We generate these explanations both in English and in the target language. 
We provide language-specific generated queries and answers in the Appendix \ref{appendix: gen_query_answer}.


\section{Experimental Setup}

In our experiments, we use the CasiMedicos dataset \cite{Agerri2023HiTZAntidoteAE} from MedExpQA benchmark \cite{ALONSO2024102938}. This dataset comprises a clinical case per question presented as multiple-choice questions, where each question includes a clinical case description, a set of potential diagnostic or treatment options, and the corresponding correct answer with explanations provided by medical doctors. The original CasiMedicos dataset provides parallel translations across four languages: English, Spanish, Italian, and French. 
The data consists of the 125 questions in the test set that we use in our experiments. 
Our test set size falls within the range of previously published medical QA research \cite{moller2020covid, gupta2024overview, gupta2025dataset}. For example, the MMLU datasets, often used for evaluation in Medical QA, are of similar size and are also included in popular initiatives, such as the Open Medical LLM Leaderboard in HuggingFace\footnote{\url{https://huggingface.co/spaces/openlifescienceai/open_medical_llm_leaderboard}}. 

Taking into account that the four languages included in the original CasiMedicos corpus represent relatively high-resource languages from the same linguistic family with extensive digital presence and computational support, we sought to extend our analysis to encompass lower-resource linguistic contexts. Thus, we additionally translated the dataset into two low-resource languages, Basque and Kazakh, using Claude-3.5-Sonnet\footnote{\url{https://www.anthropic.com/claude/sonnet}}, and then manually revised all translations with native speakers of each language. Moreover, in order to guarantee that the translations are of high quality, we evaluate all the languages through backtranslation \cite{edunov2020evaluation}. The details of this step are described in detail in Appendix \ref{appendix: backtranslation}. The results correspond to professional-grade translation quality that typically aligns with human assessments of "good" translations. The native speakers also evaluated the translations as "high-quality" with minor errors in medical abbreviations.

Such a multilingual coverage allows for a comprehensive evaluation of external knowledge integration strategies across a spectrum of languages, from well-supported higher-resource languages to underrepresented language families \cite{pfeiffer-etal-2022-lifting, chang-etal-2024-multilinguality}.

\subsection{Language of Retrieval}

We retrieve external evidence from the sources described in Section \ref{sec: sources}. Based on the multilingual availability of the sources, we aim to investigate an optimal strategy for multilingual retrieval in MMQA tasks. Therefore, we obtain results in three settings: monolingual, when retrieval is always in one language, i.e., English; multilingual, when the evidence is retrieved in the language of the question; and cross-lingual, when one part of the retrieval is done in English and the other in the language of the question \citet{liu2025xrag}. In our experiments, we retrieve the same number of 10 documents for every setting.


\subsection{Models}

To systematically evaluate the influence of different types and levels of external knowledge sources described in the preceding sections, we conducted comprehensive experiments using a diverse set of language models. Our experimental setup included the following models: 
\begin{itemize}
    \item \textbf{Qwen} \citep{bai2023qwen}: the 8B, 14B, and 32B parameter versions from Qwen 3, and the 72B parameter instruction-tuned model from Qwen 2.5.
    \item \textbf{Llama} \citep{touvron2023llama}: Llama3.1-Instruct models with 8B and 70B parameters.
    \item \textbf{Gemma} \citep{team2025gemma}: Instruction-tuned version of Gemma models with 12B and 27B parameters, and \textit{MedGemma}  \citep{sellergren2025medgemma} with 27B parameters.
\end{itemize}


The goal of the model is to analyze the question, provide contextual documents that help to find the correct answer and choose the correct option. This enables us to investigate the trade-off between information quality and quantity in retrieval-augmented generation scenarios in the medical domain. Specifically, the aim was to determine whether providing more contextual documents consistently improves performance or whether there exists an optimal balance point where additional information begins to introduce noise that degrades model accuracy. 

The contextual documents retrieved from every source differ in their content, but all of them guarantee to have the same number of documents. In the case of MedExpQA, the documents are files with reports, analysis and general definitions. Hence, the retrieved documents are more likely to describe a similar use case, but not the exact case of the input question. In LLM-generated and web-search, the contextual documents retrieved correspond to the precise answer directly addressing the search query, which may be more concise. 
All the prompts that were used for the experiments can be found in Appendix \ref{appendix:prompts}.

\section{Results and Analysis}
\label{sec: results}

This section presents the experimental results, reporting model prediction accuracy across different evaluation settings. We analyze performance trends, interpret the findings, and identify the best-performing approach.

    

\subsection{Monolingual Retrieval Results}
The experimental results from monolingual retrieval in Table \ref{tab:main_results} show significant performance variations across different retrieval methods, model sizes, and languages. Analyzing the comprehensive evaluation across six languages (English (EN), Spanish (ES), French (FR), Italian (IT), Basque (EU), and Kazakh (KK)), several key patterns emerge regarding optimal configurations for multilingual information retrieval. For all the results, we established statistical significance, as described in Appendix \ref{appendix: stats}.

\textbf{Model-wise Performance.} Intuitively, the main observation from the results is that model size significantly impacts performance, with a general trend of increasing accuracy as the number of parameters grows. The average performance across all retrieval methods and benchmarks shows a clear scaling effect. The largest models mainly exhibit the best performance across all the settings. Qwen-2.5-72B consistently outperforms the 70B version of Llama in the higher-resource languages, and underperforms in lower-resource languages, which is reflected in the high standard deviation. The performance gap reflects Qwen 2.5's poorer multilingual support compared to alternative models\footnote{\url{https://huggingface.co/Qwen/Qwen2.5-72B-Instruct}}. Although >=70B models mainly yield the highest result across all the settings, the addition of the external evidence hurts the performance in the majority of the cases, especially for Spanish, French, Italian and English.

For the rest of the models, we can see that the retrieved evidence is mainly beneficial compared to the baseline results, but they still underperform compared to the largest models.

\textbf{General Performance Trends.} The incorporation of additional external data frequently leads to performance degradation rather than improvement in high-resource languages compared to the baseline results. This trend is particularly pronounced in larger models, suggesting a negative correlation between model size and the beneficial utilization of external data sources. Nevertheless, the opposite trend is observable with the two less-resourced languages. 

Examining results by external data source, we observe the same trend: larger models consistently outperform smaller ones,  suggesting that LLMs with greater parameter counts encode sufficient domain knowledge and that retrieved information may introduce noise. 


Nevertheless, the hypothesis that larger models encode more domain-specific knowledge in their parameters is rejected by the performance of the LLM-generated evidence. Although this can be an effective strategy, it is not as powerful as external retrieval.

\textbf{Method-wise Performance.} Among all external evidence retrieval strategies evaluated, optimal performance was consistently achieved when clinical case questions were augmented with retrieved information from web-based sources, with a minor gain in results obtained with the Cohere API. Hence, in Table 2 we report the results of external evidence retrieved from Cohere API and results from Serper API are reported in Table \ref{tab:multilingual_serper_results}. 

Such performance can be attributed to the query-specific nature of web-based retrieval systems, which excel at identifying information that directly addresses the posed search query. In contrast to the medical document collection data stores, web-search engines more efficiently identify and prioritize content with high semantic relevance to the immediate query context. This distinction becomes particularly significant when compared to pre-defined knowledge bases such as MedExpQA, where document retrieval relies on similarity metrics that may identify topically related but not always directly applicable content.

\subsection{Multilingual Search Results}

In the multilingual retrieval, the observations made from the English search results are more evident. No matter the parameter size of the models, we conclude that there are no major improvements in the performance. Larger models generally perform better than smaller ones. However, with target language documents, the gains from increasing model size are less dramatic than when the retrieved documents are in English.
Moreover, the gap between higher and lower resourced languages is more evident, which could be explained primarily by the lack of sufficient information in the language in the pre-training data and the external knowledge bases. 
The influence of the performance of every LLM after adding each knowledge source in each language is shown in Appendices \ref{appendix:error_en}-\ref{appendix:error_kz}.

\subsection{Cross-lingual Search Results}

In order to find if mixing both languages of retrieval (English and the language of the question) can compensate for the lack of relevant documents in other languages, we conduct cross-lingual experiments. In this setting, we use half of the documents in English and the other half in the target language to guarantee an equal amount of retrieval across all the experimental settings. Nevertheless, as shown in Table \ref{tab:web_retrieval_comparison}, the amount of information is unequal and some document imbalance in languages takes place for Basque and Kazakh. As shown in Figure \ref{fig:retrieval_comparison}, it is evident that cross-lingual retrieval is the most optimal strategy for the low-resource languages, even when compared with monolingual English-only retrieval, and the performance accuracy becomes on par with the rest of the languages. On the contrary, the performance for Spanish, French and Italian is degraded under this setting, sometimes scoring even lower than the low-resource languages. 

\subsection{Optimal Retrieval Strategy}
Based on the data illustrated in Figure \ref{fig:retrieval_comparison}, we can conclude that the most optimal strategy for the MMQA with external evidence is to retrieve in English for high-resource languages for which the digital presence is significant, and prefer cross-lingual retrieval for the underrepresented ones. Although retrieving in the language of the question can reach significant gains for resource-rich languages, they still slightly fall short compared to the English retrieval. Similarly, for the well-known repositories such as PubMed, Wikipedia and medical textbooks, since they are in English, they are most helpful for Spanish, English, Italian and French, but not for Kazakh and Basque. Nevertheless, they underperform compared to the monolingual web retrieval, which may be motivated by the limited knowledge incorporated in these well-known medical knowledge sources. 

\section{Discussion}

\subsection{External Sources Analysis} 
Based on the results described in Section \ref{sec: results} and Appendices \ref{appendix:error_en}-\ref{appendix:error_kz}, we can conclude that smaller <30B parameter models benefit from the data provided from external sources. Nevertheless, the obvious gain is obtained for English and less frequently for French, Spanish and Italian. 

Despite these gains, overall, we see how the accuracy of the correct answer drops more frequently in the large models, indicating that either no external knowledge is beneficial or that the larger the models, the more confident they become in their internal parametric knowledge.

    

    

When it comes to answering the question of whether the information in the trusted expert data repositories, such as PubMed and Wikipedia, and the documents provided in MedExpQA, are enough, we additionally look into the retrieved sources from the web search.  In our analysis, we can see that web-search sources cover the data stores from the sources of MedExpQA. On average, 11.2\% of the information was retrieved from Wikipedia and 18.8\% of the information was retrieved from PubMed, when retrieved in English. Whereas these numbers are less than 0.03\% for the rest of the languages.




\subsection{Comparison With Other Medical Benchmarks}

To strengthen the conclusion made from our experiments, we additionally performed the same set of experiments across three other prominent medical benchmark English datasets: MedQA, PubMedQA, and MedMCQA, using web search (WS) as the external knowledge source based on the performance gain it provides in our previous experiments. The results in Table \ref{tab:other_benchmarks} demonstrate a similar pattern of improvement across all benchmarks and model sizes. 

\section{Conclusion}

This work studied how incorporating external evidence influences MMQA across different languages, retrieval methods, and model sizes. We established that web search is often the most effective external source overall; its benefits are uneven: smaller and mid-sized models usually improve with extra evidence, but larger models sometimes gain little or even degrade the performance, possibly because the provided information conflicts with what they already know. Language of retrieval also matters a lot: only English retrieval benefits the high-resource languages the most, and combining English with information in target languages is the best for low-resource settings. Curated medical sources are reliable but limited in coverage, especially outside English, while web-based evidence offers broader and more relevant information at the cost of more noise. To sum up, the results show that external knowledge does not immediately help performance; its usefulness depends on several factors such as model size, retrieval strategy, and language resources, highlighting the need for more tailored approaches in MMQA.

\section*{Limitations}

This work has several limitations. Although our conclusions are intended to be general, their applicability to other medical domains, as well as to domains outside medicine, requires further investigation. Additionally, while we found that retrieving 10 documents yielded the highest impact, our experiments were limited to document counts of up to 32. Due to computational constraints, we were unable to evaluate performance with the larger document sets and the larger models.

\bibliography{custom}

\appendix

\section{Search Query Generation} 
\label{appendix:search-query-generation}
To obtain relevant evidence via web search and LLM generation, we first generate 10 queries per question describing the clinical case using \textit{Llama-3.3-70B-Instruct}, ensuring no overlap with the models used in the main evaluation \citep{panickssery2024llm}. These queries describe the clinical case in a way that helps locate useful information for answering the question, without directly pointing to the correct answer. Consequently, the generated queries are used for the web search engine to retrieve candidate evidence passages. 

\begin{table}[h]
\section{Information availability on the web per language} 
\label{appendix: data_on_web}
\centering
\small
\begin{tabular}{l|cc|cc}
\textbf{Language} &
\multicolumn{2}{c|}{\textbf{Avg. docs / query}} &
\multicolumn{2}{c}{\textbf{Empty sources (\%)}} \\
 & Google & Cohere & Google & Cohere \\
\hline
English  & 9.24 & 4.12 & 0.00 & 1.36 \\
Spanish  & 9.29 & 3.96 & 0.00 & 1.20 \\
Italian  & 9.32 & 3.90 & 0.08 & 2.32 \\
French   & 9.32 & 3.88 & 0.00 & 1.92 \\
Basque   & 2.54 & 2.43 & 54.80 & 17.68 \\
Kazakh   & 4.17 & 2.61 & 31.36 & 22.08 \\
\end{tabular}
\caption{Retrieved document availability across languages using Google Search (via Serper) and Cohere.}
\label{tab:web_retrieval_comparison}
\end{table}

\section{Backtranslation results}
\label{appendix: backtranslation}

We backtransalte the human-written Spanish questions into each target language and back into Spanish. We evaluate the quality of backtrslations with BERTScore \cite{zhang2019bertscore}, COMET \cite{rei-etal-2020-comet}, ChrF \cite{popovic-2015-chrf} and ChrF++ \cite{popovic-2017-chrf}. BERTScore values suggest that 90-96\% of semantic content is preserved during the translation cycle, while COMET scores of 0.83-0.86 correspond to professional-grade translation quality that typically aligns with human assessments of "good" translations. 
\begin{table}[h]
\centering
\resizebox{\columnwidth}{!}{
\begin{tabular}{l|c|c|c|c}
 & BERTScore & COMET & ChrF & ChrF++ \\
\hline
ES → EN → ES & 0.9498 & 0.8538 & 83.48 & 82.05 \\
ES → IT → ES & 0.9646 & 0.8643 & 88.45 & 87.58 \\
ES → FR → ES & 0.9513 & 0.8564 & 84.61 & 83.23 \\
ES → KZ → ES & 0.9013 & 0.8349 & 71.00 & 68.75 \\
ES → EU → ES & 0.9437 & 0.8549 & 82.22 & 80.70 \\
\end{tabular}
}
\caption{Results of backtranslation from Spanish (original language of the dataset) to each language.}
\label{tab:backtranslation}
\end{table}

\section{Statistical Significance}
\label{appendix: stats}
To establish whether performance differences between the baseline (without retrieval) and each retrieval strategy are statistically significant, we performed chi-square tests of independence on model prediction outcomes. For each language and model, we constructed a 2×2 contingency table comparing correct vs. incorrect predictions under the baseline and retrieval-augmented conditions. All the comparisons produced p-values < 0.001. Given the large and consistent effect sizes observed across languages and model scales, we did not observe borderline cases sensitive to the choice of significance threshold.

\begin{figure*}[t]
\section{Performance of retrieving different numbers of documents using MedExpQA as context.}
\label{appendix:doc_count_figures}
  \centering
  \begin{tabular}{cc}
    \includegraphics[width=0.45\textwidth]{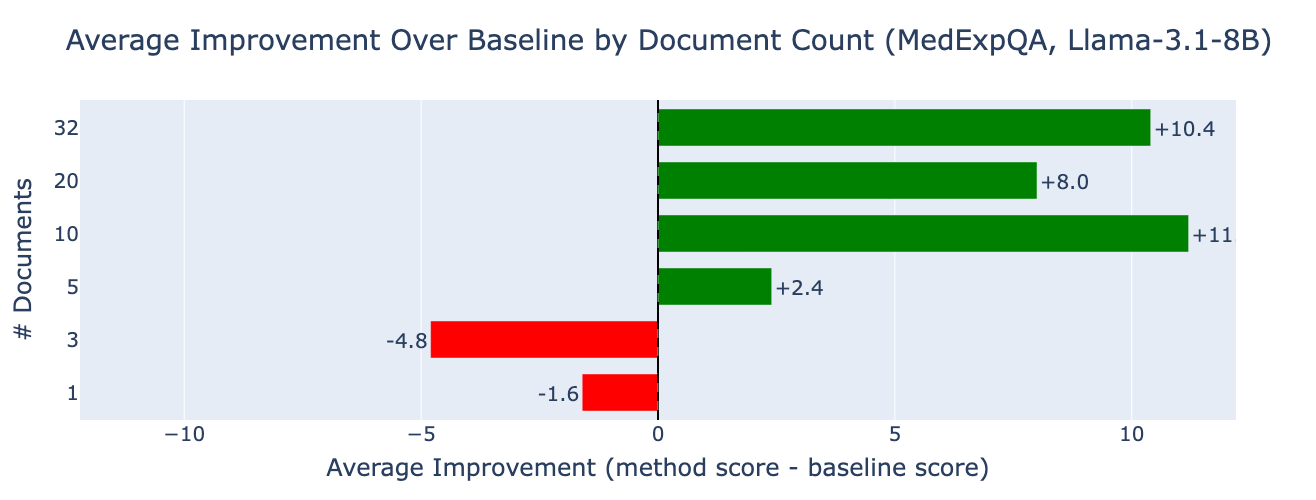} &
    \includegraphics[width=0.45\textwidth]{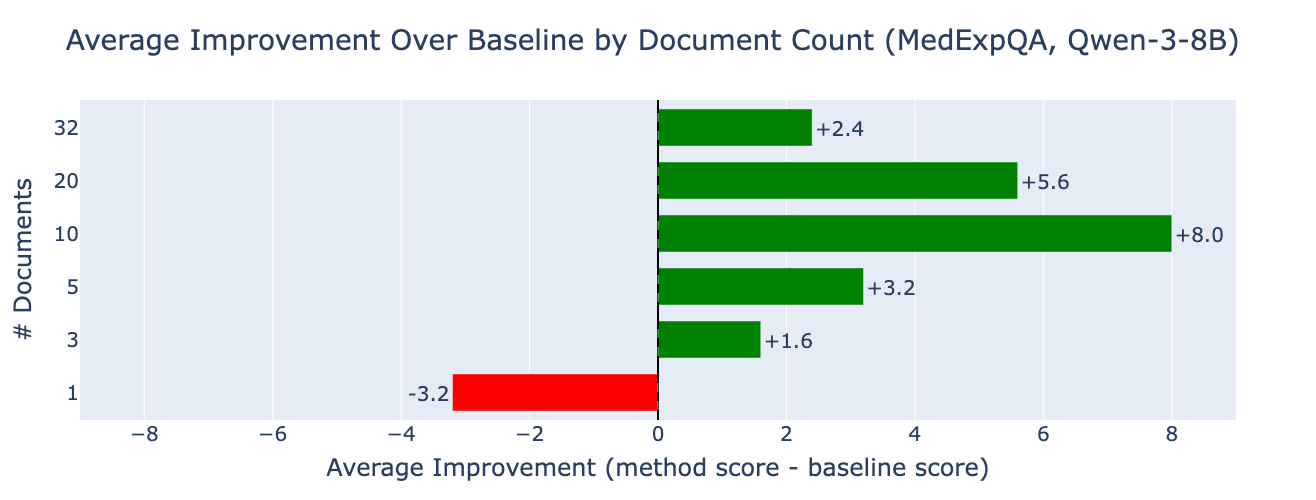} \\
    \includegraphics[width=0.45\textwidth]{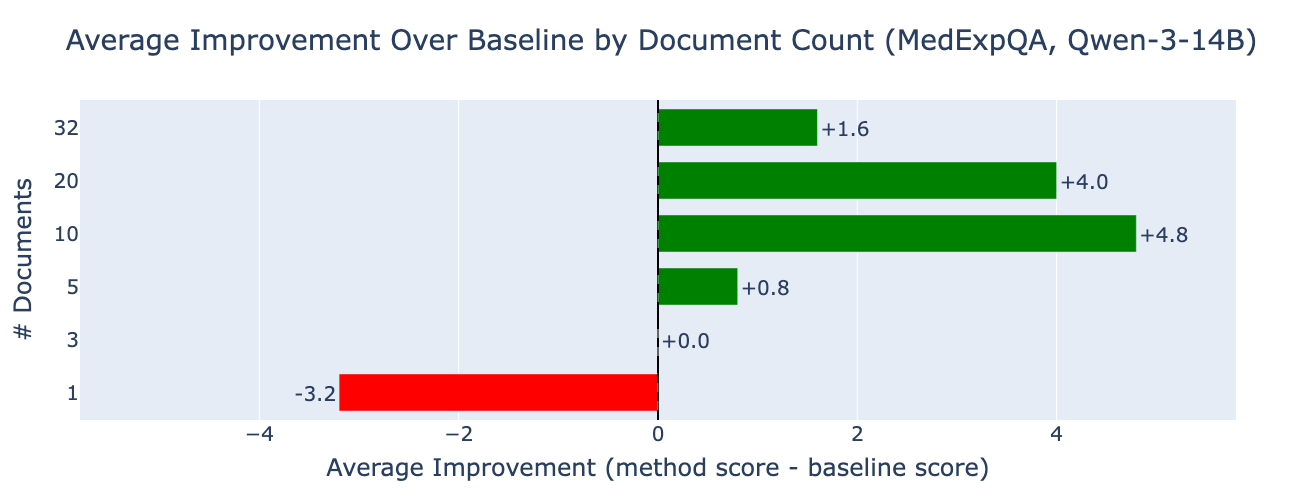} &
    \includegraphics[width=0.45\textwidth]{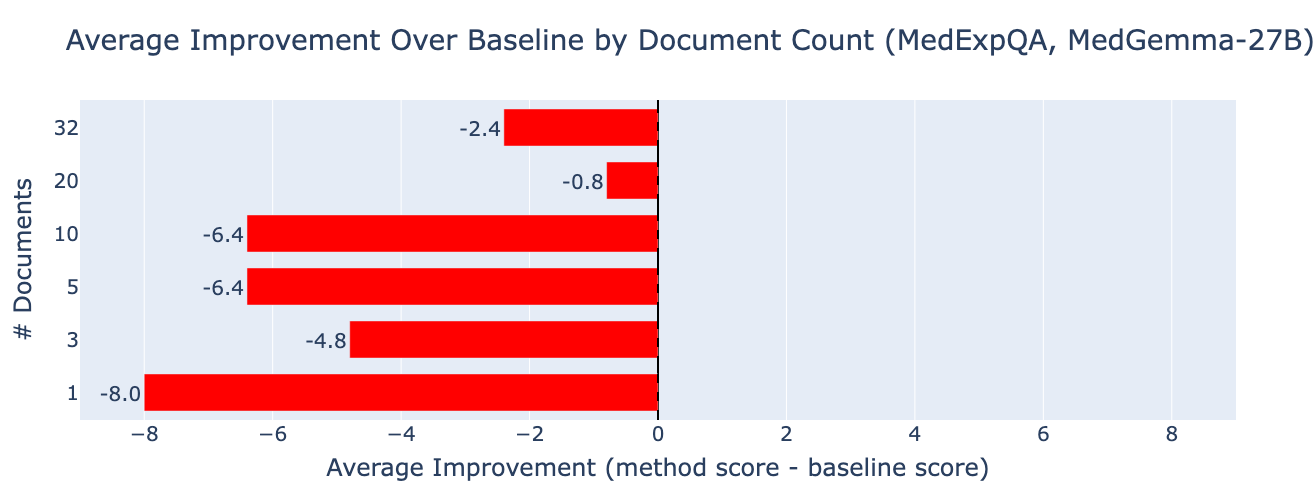} 
  \end{tabular}
  \caption{Performance of different models under different document counts from MedExpQA.}
  \label{fig:medexpqa-doccount-main}
\end{figure*}


\section{Multilingual prompts for multiple choice question-answering}
\label{appendix:prompts}

\newtcolorbox{evidencebox}[1][]{
  enhanced,
  breakable,
  colback=white,
  colframe=black,
  boxrule=2pt,
  arc=8pt,
  left=10pt,
  right=10pt,
  top=10pt,
  bottom=10pt,
  fontupper=\normalsize,
  #1
}

Prompt used for English experiments:
\begin{evidencebox}
\textbf{ You are an expert medical AI assistant tasked with answering multiple-choice medical questions for research purposes. Your goal is to analyze the given clinical case, consider the provided options, and select the correct answer. You are given a set of relevant documents that will help you to find the correct answer, analyze them and answer the question.} \newline

Here are the documents:  \newline
\textbf{\textit{[set of retrieved documents]}} \newline \newline
Here is the clinical case: \newline  
\textbf{\textit{[clinical case question]}} \newline \newline
The options:  \newline
\textbf{\textit{[set of options]}} \newline 
\end{evidencebox}

Prompt used for Spanish experiments:
\begin{evidencebox}
\textbf{Eres un asistente médico experto encargado de responder preguntas médicas de opción múltiple con fines de investigación. Tu objetivo es analizar el caso clínico dado, considerar las opciones proporcionadas y seleccionar la respuesta correcta. Se le proporciona un conjunto de documentos relevantes que le ayudarán a encontrar la respuesta correcta, analizarlos y responder la pregunta.}
\newline

Aquí están los documentos: \newline
\textbf{\textit{[set of retrieved documents]}} \newline \newline
Aquí está el caso clínico: \newline
\textbf{\textit{[clinical case question]}} \newline \newline
Aquí están las posibles opciones: \newline
\textbf{\textit{[clinical case question]}} \newline
\end{evidencebox}

Prompt used for French experiments:
\begin{evidencebox}
\textbf{Vous êtes un assistant médical qualifié chargé de répondre à des questions médicales à choix multiples dans le cadre d'une recherche. Votre objectif est d'analyser le cas clinique présenté, d'examiner les options proposées et de sélectionner la bonne réponse. Vous disposez d'un ensemble de documents pertinents qui vous aideront à trouver la bonne réponse; analysez-les et répondez à la question.} \newline

Voici les documents: \newline
\textbf{\textit{[set of retrieved documents]}} \newline \newline
Voici le cas clinique: \newline
\textbf{\textit{[clinical case question]}} \newline \newline
Voici les options possibles: \newline
\textbf{\textit{[set of options]}} \newline
\end{evidencebox}

Prompt used for Italian experiments:
\begin{evidencebox}
\textbf{[IT] Sei un assistente medico qualificato incaricato di rispondere a domande mediche a risposta multipla per scopi di ricerca. Il tuo obiettivo è analizzare il caso clinico fornito, considerare le opzioni fornite e selezionare la risposta corretta. Ti verrà fornita una serie di documenti pertinenti che ti aiuteranno a trovare la risposta corretta, ad analizzarli e a rispondere alla domanda.} \newline

[IT] Ecco i documenti: \newline
\textbf{\textit{[set of retrieved documents]}} \newline \newline
[IT] Ecco il caso clinico: \newline
\textbf{\textit{[clinical case question]}} \newline \newline
[IT] Ecco le opzioni possibili: \newline
\textbf{\textit{[set of options]}} \newline
\end{evidencebox}

Prompt used for Basque experiments:
\begin{evidencebox}
\textbf{[EU] Medikuntzako laguntzaile trebea zara, ikerketa-helburuetarako aukera anitzeko galdera medikoei erantzuteko ardura duena. Zure helburua kasu klinikoa aztertzea, emandako aukerak kontuan hartzea eta erantzun zuzena hautatzea da. Erantzun zuzena aurkitzen, aztertzen eta galderari erantzuten lagunduko dizuten dokumentu multzo garrantzitsu bat ematen zaizu.}
\newline
[EU] Hemen daude dokumentuak: \newline
\textbf{\textit{[set of retrieved documents]}} 
\newline
[EU] Hona hemen kasu klinikoa: \newline
\textbf{\textit{[clinical case question]}}
\newline

[EU] Hona hemen aukera posibleak: \newline
\textbf{\textit{[set of options]}}

\end{evidencebox}

\section{Prompts used for query generation}


\begin{evidencebox}
\textbf{You are given a question describing a clinical case. You have to generate a list of 10 queries for a web-search that will help to find more information to help to find the correct answer to the clinical case. Provide your answer in Python list format. Only in the list form, your answer should start with [ and end with ]. Do not give any other answer. Do not give explanations.} 
\newline
\newline
Clinical case: 

\textit{{\textbf{[Clinical case topic]}}}
\newline
\newline
Questions:

\textit{{\textbf{ [Clinical case question]}}}

\end{evidencebox}

\begin{table*}[t]
\section{Generated queries and explanations}
\label{appendix: gen_query_answer}
\centering
\begin{tabular}{p{0.18\textwidth} | p{0.78\textwidth}}
\textbf{Generated query} & \textbf{Generated answer} \\ \hline

[EN] What are the common complications of acute myocardial infarction? &
Common complications of acute myocardial infarction include heart failure,
arrhythmias, and cardiac rupture. Other complications are pericarditis,
mitral regurgitation, ventricular septal defect, cardiogenic shock, and
pulmonary edema. Thromboembolism and Dressler syndrome can also occur. \\
\hline
[ES] ¿Cuáles son las complicaciones comunes del infarto agudo de miocardio? &
Las complicaciones comunes del infarto agudo de miocardio incluyen insuficiencia
cardíaca, arritmias, choque cardiogénico, rotura ventricular y muerte súbita.
La insuficiencia cardíaca es frecuente y puede resultar de la pérdida de función
del músculo cardíaco. Las arritmias pueden ser mortales si no se tratan
rápidamente. El choque cardiogénico ocurre cuando el corazón no bombea
suficiente sangre. La rotura ventricular es rara pero grave. \\
\hline
[IT] Quali sono le complicanze più comuni dell'infarto miocardico acuto? & Le complicazioni comuni dell'infarto miocardico acuto includono insufficienza cardiaca, aritmie, morte improvvisa, rottura del setto interventricolare e
 tromboembolismo. L'insufficienza cardiaca è una delle complicazioni più comuni e può essere causata da una riduzione della funzione contrattile del cuore. Le aritmie, come la fibrillazione ventricolare e la tachicardia ventricolare, possono verificarsi a causa della lesione del miocardio. La morte improvvisa può verificarsi a causa di aritmie gravi o di una grave insufficienza cardiaca. La rottura del setto interventricolare e il tromboembolismo sono complicazioni meno comuni, ma possono essere gravi e potenzialmente letali. \\
 \hline
 [FR] Quelles sont les complications courantes d'un infarctus aigu du myocarde?  &  Les complications courantes de l'infarctus aigu du myocarde comprennent l'insuffisance cardiaque, l'arrêt cardiaque, la fibrillation ventriculaire, la rupture du muscle cardiaque et l'embolie pulmonaire. L'insuffisance cardiaque est une complication majeure qui peut survenir immédiatement ou des semaines après l'infarctus. Les autres complications incluent la tamponnade cardiaque, la thrombose veineuse profonde et la pneumonie. Les complications peuvent varier en fonction de la gravité de l'infarctus et de la rapidité du traitement. Un traitement précoce et des soins appropriés peuvent contribuer à prévenir ou à atténuer ces complications. \\
 \hline
 [EU] Zeintzuk dira miokardioko infartu akutuaren konplikazio ohikoenak? & Miokardio infartu akutuaren konplikazio arruntenak bihotz-gutxiegitasuna, takikardia bentrikularrak, blokeo aurikulobentrikularra eta arritmia kardiakoa dira. Konplikazio hauek heriotza kardiakoa eragin dezakete. Bihotz-gutxiegitasunak bihotzak odola ponpatzeko duen gaitasuna murrizten du. Takikardia bentrikularrak bihotzaren taupada azkarra eragiten du. Arritmia kardiakoak bihotzaren erritmoa desordenatzen du.  Bihotzeko aneurisma ere gerta daiteke.  Konplikazio hauek tratatu eta prebenitzeko, medikuak sendagaiak, kirurgia eta beste tratamendu batzuk erabiltzen ditu.  Konplikazio hauek tratatzeko, ospitalean sartzea eta jarraipena behar da. \\
 \hline

\end{tabular}
\caption{Example of generated queries and generated answers to those queries for each language.}
\label{tab:query_gen}
\end{table*}

\begin{table*}[h]
\scriptsize
\setlength{\tabcolsep}{0.9pt} 
\renewcommand{\arraystretch}{1} 

\section{Monolingual evidence results with Serper API}

\label{appendix:multilingual_serper}
\begin{center}
\begin{tabular}{c|cc|cc|ccc|cc>{\columncolor{lightgreen}}c}
\rowcolor{headercolor}
\textbf{Params} & \multicolumn{2}{c|}{\textbf{8B}} & \multicolumn{2}{c|}{\textbf{12B-14B}} & \multicolumn{3}{c|}{\textbf{27-32B}} & \multicolumn{2}{c|}{\textbf{>=70B}} & \textbf{Avg (Std.)} \\
\rowcolor{headercolor}
\hline
Models & LLaMA & Qwen & Gemma & Qwen & Gemma & MedGemma & Qwen & LlaMA & Qwen & \\
\rowcolor{headercolor}
& \textbf{8B} & \textbf{8B} & \textbf{12B} & \textbf{14B} & \textbf{27B} & \textbf{27B} & \textbf{32B} & \textbf{70B} & \textbf{72B} & \\
\hline
\rowcolor{darkgreen}
\multicolumn{11}{c}{\textit{\textbf{Web-search (Serper)}}} \\
ES & 68.0 & 68.8 & 72.8 & 77.6 & 71.2 & 76.8 & 80.0 & 79.2 & 76.8 & 74.58 (4.23) \\
FR & 68.0 & 69.6 & 67.2 & 76.0 & 69.6 & 75.2 & 78.4 & 80.8 & 77.6 & 73.6 (4.76) \\
IT & 65.6 & 69.6 & 69.6 & 80.0 & 69.6 & 76.0 & 76.8 & 79.2 & 80.0 & 74.04 (5.16) \\
EU & 55.2 & 58.4 & 66.4 & 66.4 & 69.6 & 76.8 & 68.0 & 76.0 & 63.2 & 66.67 (6.77) \\
KK & 54.4 & 60.0 & 61.6 & 58.4 & 66.4 & 74.4 & 68.0 & 70.4 & 60.8 & 3.82 (6.02)  \\
\end{tabular}

\caption{Performance comparison across different monolingual English retrieval methods using Serper API.}  
\label{tab:multilingual_serper_results}

\end{center}
\end{table*}

\begin{table*}[h]
\scriptsize
\setlength{\tabcolsep}{0.8pt} 
\renewcommand{\arraystretch}{1} 
\section{Multilingual evidence results}
\label{appendix:multilingual}
\begin{center}
\begin{tabular}{c|cc|cc|ccc|cc>{\columncolor{lightgreen}}c}
\rowcolor{headercolor}
\textbf{Params} & \multicolumn{2}{c|}{\textbf{8B}} & \multicolumn{2}{c|}{\textbf{12B-14B}} & \multicolumn{3}{c|}{\textbf{27-32B}} & \multicolumn{2}{c|}{\textbf{>=70B}} & \textbf{Avg (Std.)} \\
\rowcolor{headercolor}
\hline
Models & LLaMA & Qwen & Gemma & Qwen & Gemma & MedGemma & Qwen & LlaMA & Qwen & \\
\rowcolor{headercolor}
& \textbf{8B} & \textbf{8B} & \textbf{12B} & \textbf{14B} & \textbf{27B} & \textbf{27B} & \textbf{32B} & \textbf{70B} & \textbf{72B} & \\
\hline
\rowcolor{darkgreen}
\multicolumn{11}{c}{\textbf{LLM's parametric knowledge}} \\
ES & 58.4 & 70.4 & 71.2 & 72.8 & 71.2 & 78.4 & 75.2 & 73.6 & 76.0 & 71.91 (5.37) \\
FR & 62.4 & 69.6 & 65.6 & 71.2 & 71.2 & 75.2 & 75.2 & 79.2 & 78.4 & 72.0 (5.31) \\
IT & 64.8 & 72.8 & 68.8 & 75.2 & 73.6 & 78.4 & 77.6 & 78.4 & 80.8 & 74.49 (4.84) \\
EU & 52.0 & 60.8 & 63.2 & 61.6 & 71.2 & 74.4 & 65.6 & 72.8 & 59.2 & 64.53 (6.84) \\
KK & 46.4 & 53.6 & 60.0 & 57.6 & 64.0 & 67.2 & 64.0 & 68.8 & 65.6 & 60.8 (6.82)  \\
\hline
\rowcolor{darkgreen}
\multicolumn{11}{c}{\textbf{ \textit{Web Search (Cohere)}}} \\
ES & 64.0 & 74.4 & 72.8 & 75.2 & 78.4 & 79.2 & 79.2 & 80.0 & 77.6 & 75.64 (4.73) \\
FR & 64.8 & 73.6 & 72.0 & 72.0 & 73.6 & 78.4 & 78.4 & 79.2 & 75.2 & 74.13 (4.22) \\
IT & 66.4 & 72.8 & 71.2 & 76.8 & 78.4 & 75.2 & 79.2 & 80.0 & 78.4 & 75.38 (4.23) \\
EU & 47.2 & 52.0 & 61.6 & 56.8 & 63.2 & 69.6 & 62.4 & 68.0 & 59.2 & 60.0 (6.78) \\
KK & 48.0 & 32.0 & 50.4 & 35.2 & 59.2 & 61.6 & 35.2 & 70.4  & 41.6 & 40.36 (17.38) \\
\rowcolor{darkgreen}
\multicolumn{11}{c}{\textbf{ \textit{Web Search (Serper)}}} \\
ES & 60.8 & 72.8 & 72.8 & 77.6 & 78.4 & 80.0 & 78.4 & 80.0 & 79.2 & 75.56 (5.83) \\
FR & 64.8 & 71.2 & 70.4 & 74.4 & 76.8 & 77.6 & 76.8 & 76.0 & 79.2 & 74.13 (4.28) \\
IT & 63.2 & 68.0 & 70.4 & 72.8 & 78.4 & 73.6 & 80.0 & 81.6 & 74.4 & 73.60 (5.57) \\
EU & 38.4 & 45.6 & 57.6 & 53.6 & 68.0 & 70.4 & 62.0 & 69.6 & 59.2 & 58.27 (10.34) \\
KK & 39.2 & 44.8 & 58.4 & 51.2 & 64.0 & 64.0 & 56.0 & 68.8 & 58.4 & 56.09 (9.03) \\
\end{tabular}

\caption{Performance comparison across different multilingual retrieval methods, model sizes and languages, meaning the retrieved documents are in the language of the question. The models are grouped by the parameter size.}  
\label{tab:multilingual_results}

\end{center}
\end{table*}

\begin{table*}[h]
\section{Cross-lingual evidence results}
\label{appendix:crosslingual}

\scriptsize 
\setlength{\tabcolsep}{0.9pt} 
\renewcommand{\arraystretch}{1} 
\begin{center}
\begin{tabular}{c|cc|cc|ccc|cc|>{\columncolor{lightgreen}}c}
\rowcolor{headercolor}
\textbf{Params} & \multicolumn{2}{c|}{\textbf{8B}} & \multicolumn{2}{c|}{\textbf{12B-14B}} & \multicolumn{3}{c|}{\textbf{27-32B}} & \multicolumn{2}{c|}{\textbf{>=70B}} & \textbf{Avg (Std.)} \\
\rowcolor{headercolor}
\hline
Models & LLaMA & Qwen & Gemma & Qwen & Gemma & MedGemma & Qwen & LlaMA & Qwen & \\
\rowcolor{headercolor}
& \textbf{8B} & \textbf{8B} & \textbf{12B} & \textbf{14B} & \textbf{27B} & \textbf{27B} & \textbf{32B} & \textbf{70B} & \textbf{72B}  & \\
\hline
\rowcolor{blue}
\multicolumn{11}{c}{\textbf{LLM's parametric knowledge}} \\
ES & 61.6 & 71.2 & 60.8 & 74.4 & 68.8 & 76.0 & 75.2 & 80.0 & 76.8 & 71.64 (6.35) \\
FR & 65.6 & 66.4 & 66.4 & 69.6 & 71.2 & 68.8 & 74.4 & 76.8 & 80.8 & 71.11 (4.94)  \\
IT & 65.6 & 70.4 & 65.6 & 72.8 & 72.0 & 72.8 & 77.6 & 72.8 & 79.2 & 72.09 (4.34) \\
EU & 60.0 & 63.2 & 59.2 & 69.6 & 72.0 & 76.8 & 77.6 & 75.2 & 80.0 & 70.4 (7.43) \\
KK & 65.6 & 68.8 & 67.2 & 71.2 & 71.2 & 71.2 & 72.8 & 76.8 & 76.8 & 71.29 (3.63)  \\
\hline
\rowcolor{blue}
\multicolumn{11}{c}{\textbf{ \textit{Web Search (Cohere)}}} \\
ES & 65.6 & 75.2 & 72.0 & 79.2 & 77.6 & 77.6 & 80.8 & 77.6 & 80.0 & 76.18 (4.49) \\
FR & 72.0 & 78.4 & 71.2 & 76.0 & 72.0 & 73.6 & 84.0 & 79.2 & 80.8 & 76.36 (4.25) \\
IT & 66.4 & 79.2 & 69.6 & 76.8 & 74.4 & 76.8 & 77.6 & 78.4 & 84.0 & 75.91 (4.94) \\
EU & 62.4 & 71.2 & 75.2 & 80.0 & 75.2 & 80.8 & 76.8 & 80.8 & 81.6 & 76.00 (5.81) \\
KK & 67.2 & 75.2 & 70.4 & 75.2 & 75.2 & 78.4 & 75.2 & 81.6 & 83.2 & 75.73 (4.71) \\

\hline
\rowcolor{blue}
\multicolumn{11}{c}{\textbf{ \textit{Web Search (Serper)}}} \\
ES & 57.6 & 75.2 & 62.4 & 74.4 & 72.0 & 72.0 & 78.4 & 76.8 & 80.8 & 72.18 (7.12) \\
FR & 63.2 & 70.4 & 69.6 & 75.2 & 76.8 & 76.8 & 79.2 & 79.2 & 79.2 & 74.4 (5.23) \\
IT & 70.4 & 70.4 & 76.8 & 75.2 & 76.8 & 76.8 & 77.6 & 79.2 & 77.6 & 75.64 (2.97) \\
EU & 65.6 & 70.4 & 64.8 & 69.6 & 73.6 & 77.6 & 79.2 & 73.6 & 77.6 & 72.44 (4.93) \\
KK & 68.8 & 70.4 & 68.8 & 68.0 & 73.6 & 72.8 & 74.4 & 78.4 & 78.4 & 72.62 (3.75) \\

\end{tabular}

\caption{Performance comparison across different cross-lingual retrieval methods, model sizes and languages, meaning the retrieved documents are in the language of the question and in English. The models are grouped by the parameter size.}  
\label{tab:crosslingual_results}

\end{center}
\end{table*}

\begin{table*}[h]
\centering
\section{Web-search Results For Other Methods}
\label{appendix: pther_benchmarks}
\begin{tabular}{lcccccc}
\toprule
& \multicolumn{2}{c}{\textbf{MedQA}} & \multicolumn{2}{c}{\textbf{MedMCQA}} & \multicolumn{2}{c}{\textbf{PubMedQA}} \\
\cmidrule(lr){2-3} \cmidrule(lr){4-5} \cmidrule(lr){6-7}
\textbf{Model} & \textbf{Base} & \textbf{+WS} & \textbf{Base} & \textbf{+WS} & \textbf{Base} & \textbf{+WS} \\
\midrule
\multicolumn{7}{l}{\textit{8-14B}} \\
\midrule
Llama3.1-8B    & 56.17 & 64.34 & 55.10 & 64.64 & 70.4 & 74.6 \\
Qwen3-8B       & 58.44 & 67.71 & 56.42 & 65.50 & 72.4 & 71.0 \\
Gemma3-12B     & 56.01 & 61.04 & 51.23 & 54.75 & 66.0 & 76.4 \\
Qwen3-14B      & 63.01 & 69.59 & 56.89 & 64.79 & 72.0 & 77.6 \\
\midrule
\multicolumn{7}{l}{\textit{27-32B}} \\
\midrule
Gemma3-27B     & 61.74 & 65.75 & 59.72 & 63.35 & 70.6 & 68.0 \\
MedGemma-27B   & 67.95 & 70.46 & 61.85 & 66.91 & 70.0 & 73.8 \\
Qwen3-32B      & 63.79 & 73.29 & 64.79 & 56.99 & 74.4 & 76.8 \\
\midrule
\multicolumn{7}{l}{\textit{70-72B}} \\
\midrule
Llama3.1-70B   & 72.82 & 76.12 & 69.73 & 74.09 & 77.6 & 79.6 \\
Qwen2.5-72B    & 70.93 & 74.16 & 66.82 & 72.84 & 76.7 & 78.4 \\
\bottomrule
\end{tabular}
\caption{Performance comparison of language models with and without retrieved external data (WS) across English Medical QA benchmarks. Across the 8-14B parameter models, we observe substantial gains: Llama3.1-8B shows improvements of 8.17, 9.54, and 4.2 points on MedQA, MedMCQA, and PubMedQA, respectively, while Qwen3-14B demonstrates gains of 6.58, 7.90, and 5.6 points. The 27-32B models exhibit similar trends, with Qwen3-32B achieving a notable 9.5-point improvement on MedQA, and even the medical-specific MedGemma-27B showing consistent gains despite its domain fine-tuning.}
\label{tab:other_benchmarks}
\end{table*}

\begin{figure*}
    \centering
    \section{Error rate by every external knowledge source and LLM in English}
    \label{appendix:error_en}
    \includegraphics[width=1\linewidth]{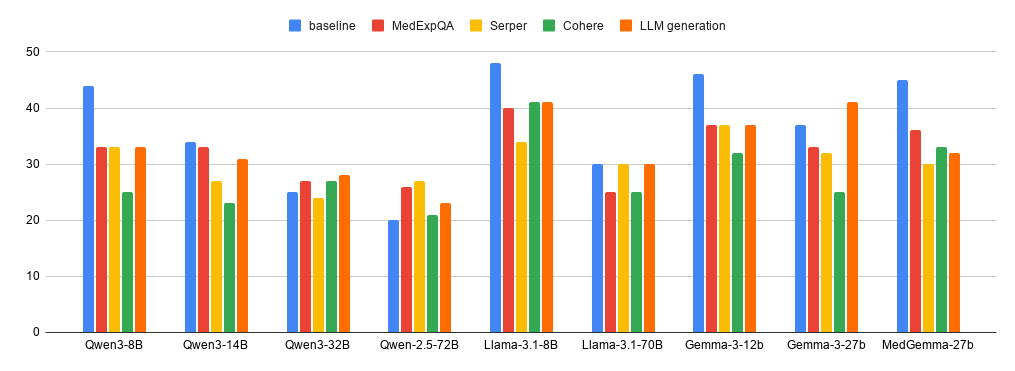}
    \caption{Error rate by every external knowledge source and LLM in English}
    \label{fig:error_en}
\end{figure*}

\begin{figure*}
    \centering
    \section{Error rate by every external knowledge source and LLM in Spanish}
    \includegraphics[width=1\linewidth]{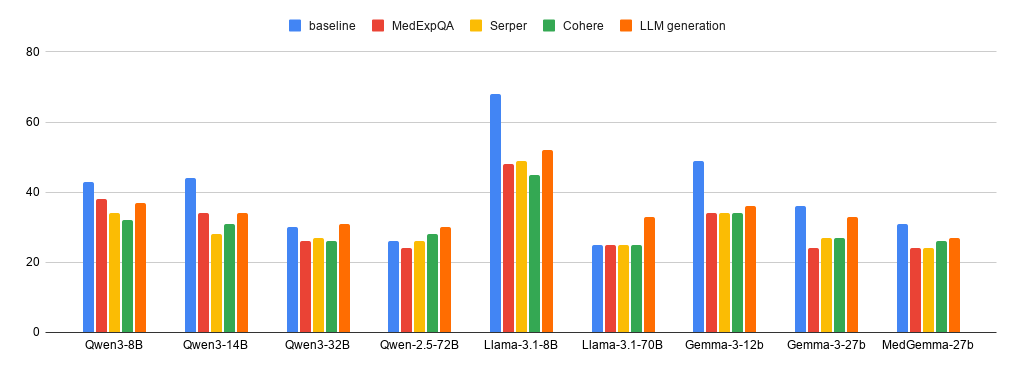}
    \caption{Error rate by every external knowledge source and LLM in Spanish}
    \label{fig:error_es}
\end{figure*}

\begin{figure*}
    \centering
    \section{Error rate by every external knowledge source and LLM in Italian}
    \includegraphics[width=1\linewidth]{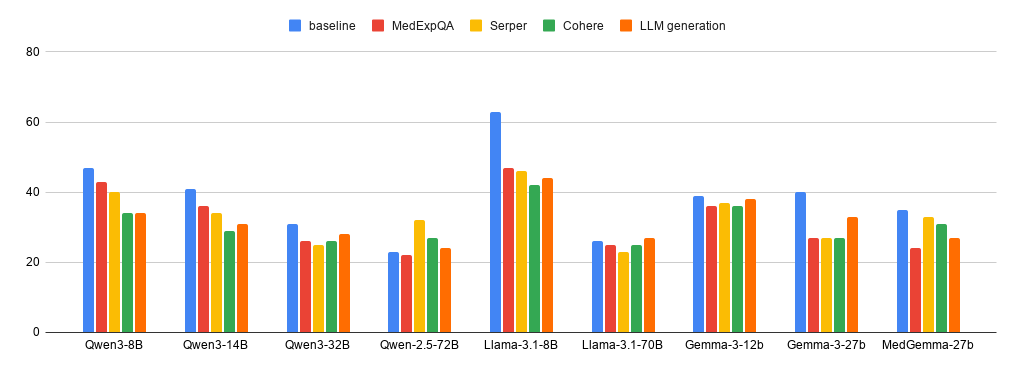}
    \caption{Error rate by every external knowledge source and LLM in Italian}
    \label{fig:error_it}
\end{figure*}

\begin{figure*}
    \centering
    \section{Error rate by every external knowledge source and LLM in French}
    \includegraphics[width=1\linewidth]{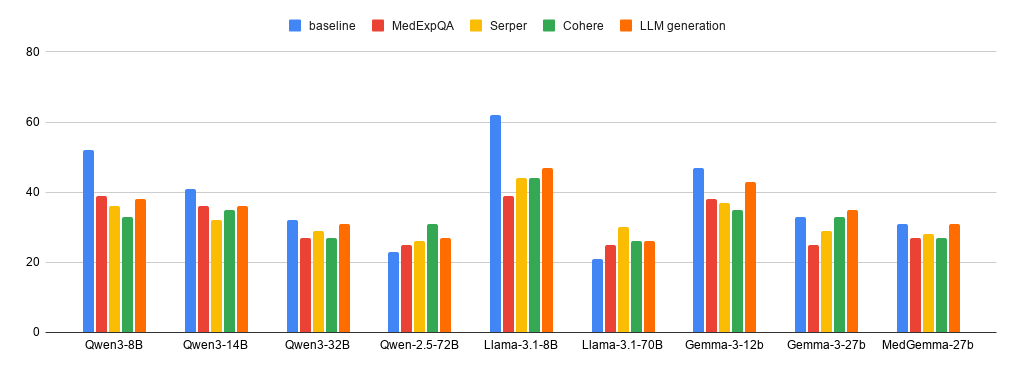}
    \caption{Error rate by every external knowledge source and LLM in French}
    \label{fig:error_fr}
\end{figure*}

\begin{figure*}
    \centering
    \section{Error rate by every external knowledge source and LLM in Basque}
    \includegraphics[width=1\linewidth]{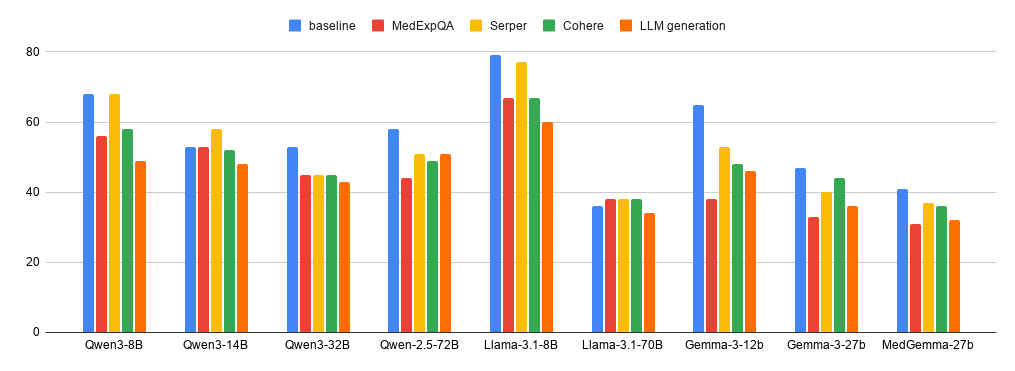}
    \caption{Error rate by every external knowledge source and LLM in Basque}
    \label{fig:error_eu}
\end{figure*}

\begin{figure*}
    \centering
    \section{Error rate by every external knowledge source and LLM in Kazakh}
    \label{appendix:error_kz}
    \includegraphics[width=1\linewidth]{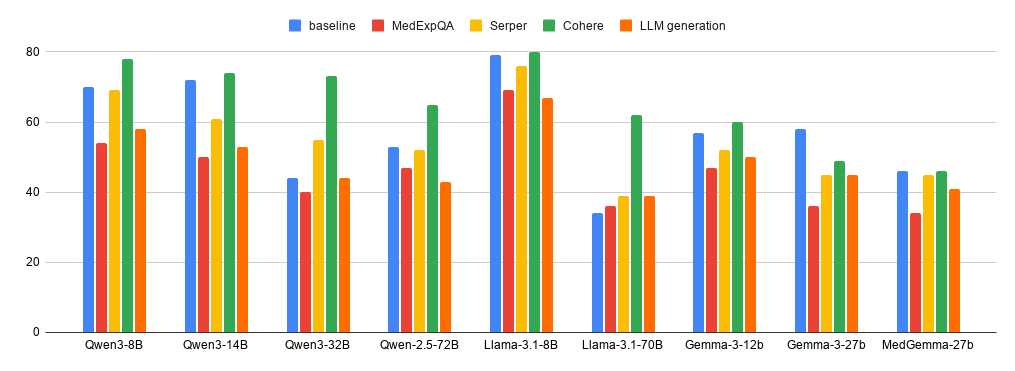}
    \caption{Error rate by every external knowledge source and LLM in Kazakh}
    \label{fig:error_kz}
\end{figure*}

\end{document}